\documentclass{article}
\usepackage{ijcai25}
\usepackage{multirow}
\usepackage{times}
\usepackage{soul}
\usepackage{url}
\usepackage[hidelinks]{hyperref}
\usepackage[utf8]{inputenc}
\usepackage[small]{caption}
\usepackage{graphicx}
\usepackage{amsmath}
\usepackage{amsthm}
\usepackage{booktabs}
\usepackage{algorithm}
\usepackage{algorithmic}
\usepackage[switch]{lineno}
\usepackage{hyperref}
\usepackage{xcolor}
\usepackage{tikz}
\usepackage{pgfplots}

\title{Tsururu: A Python-based Time Series Forecasting Strategies Library}

\author{
Alina Kostromina$^1$
\and
Kseniia Kuvshinova$^{1,3}$\and
Aleksandr Yugay$^{2,4}$\and
Andrey Savchenko$^{1,5}$\and \\
Dmitry Simakov$^1$\\
\affiliations
$^1$Sber AI Lab, Moscow, Russia\\
$^2$Sber, Moscow, Russia\\
$^3$Skoltech, Moscow, Russia\\
$^4$MIPT, Moscow, Russia\\
$^5$ISP RAS Research Center for Trusted Artificial Intelligence, Moscow, Russia\\
\emails
alina.kostromina@gmail.com
}

\begin{document}

\maketitle
\begin{abstract}
While current time series research focuses on developing new models, crucial questions of selecting an optimal approach for training such models are underexplored. Tsururu, a Python library introduced in this paper, bridges SoTA research and industry by enabling flexible combinations of global and multivariate approaches and multi-step-ahead forecasting strategies. It also enables seamless integration with various forecasting models. Available at \url{https://github.com/sb-ai-lab/tsururu}.
\end{abstract}
\section{Introduction}

A fundamental task in time series analysis is forecasting, which involves predicting future values $X_{t+1:H}$ over a horizon $H$ at timestep $t$, given the historical
data $X_{1:t}$ and known covariates $Z_{1:H}$ for all time points \cite{salinas2020deepar,kim2024comprehensive}. Later in this paper, we consider multivariate data comprising multiple time series within the dataset.

To advance forecasting capabilities, various libraries have been developed to benchmark state-of-the-art (SoTA) models, including tslib \cite{wang2024deep}, neuralforecast \cite{olivares2022library_neuralforecast}, uni2ts \cite{aksu2024gift} (for in-context learning models), pytorch-forecasting \cite{pytorch-forecasting} (has earlier deep learning models), BasicTS \cite{shao2024exploring}.
However, they are suboptimal for real-world scenarios. They often rely on fixed forecasting strategies, offer constrained support for exogenous variables, and face difficulties in the usage of custom datasets. 
These factors limit their suitability for industrial scenarios. A potential solution is a time series library that seamlessly handles diverse datasets, including non-aligned ones or those with exogenous features.

Indeed, some of the issues above are partially covered in existing practically oriented time series libraries. Almost all such libraries use global approach \cite{januschowski2020criteria,montero2021principles} (useful for non-aligned series) and support exogenous variables: Darts \cite{JMLR:v23:21-1177}, sktime \cite{franz_kiraly_2025_14808427}, gluonts \cite{gluonts_jmlr},  tslearn \cite{JMLR:v21:20-091}, skforecast \cite{skforecast}, AutoTS \cite{AutoTS} (global for aligned time series only), ETNA \cite{etna}, mlforecast \cite{mlforecast}. However, some of them could not use exogenous features as input, like tslearn \cite{JMLR:v21:20-091}, and tsspiral \cite{tspiral} or work only in the multivariate setting, such as tsai \cite{tsai}.

Nevertheless, the full predictive potential of time series models remains unexplored, as the impact of forecasting strategies \cite{taieb2014machine} has received limited attention. Classical time series forecasting is based either on a recursive strategy \cite{gustin2018forecasting} or a multi-input-multi-output (MIMO) strategy \cite{bontempi2008long,bontempi2011conditionally,kline2004methods}. Recent research claimed that the question of which strategy to use is still open \cite{green2024time}. 
Notably, most state-of-the-art neural networks are trained using the MIMO strategy. However, incorporating additional forecasting strategies demonstrates that MIMO is not always optimal (see Section~\ref{section:case}). 
Only Darts, sktime, skforecast and tspiral have at least three strategies. Nevertheless, they offer only a narrow pool of preprocessing methods. For example, none of them is integrated with the LastKnownNormalizer — subtraction or division on the last element in available history. Our experiments show that this rarely used preprocessing enhances forecasting performance significantly (see Section~\ref{section:case}).

\begin{figure*}[htbp]
    \centering
    \includegraphics[width=\textwidth]{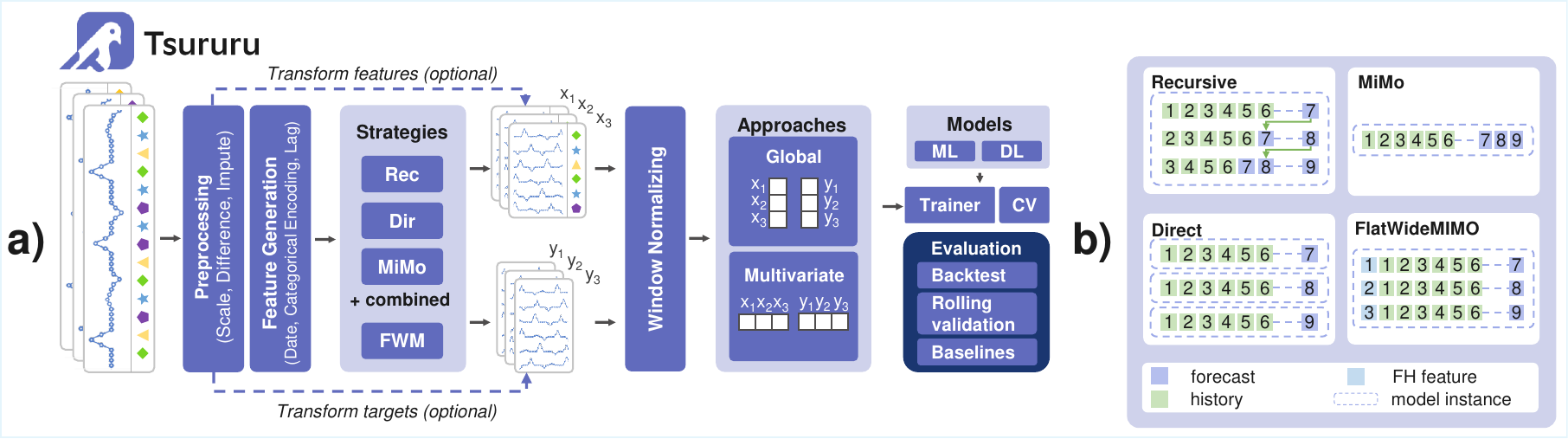}
    \caption{a) The architecture of the proposed Tsururu framework. Most existing libraries typically support either (1) global and multivariate forecasting or (2) multiple forecasting strategies, but not both. Tsururu enables the exploration of all possible combinations of available ML and DL models with various forecasting strategies, preprocessing techniques, and the global/multivariate setting.
    b) The inference stage of the forecasting process for each strategy. FH stands for forecasting horizon.
    }
    \label{fig:dataflow}
\end{figure*}

Given the weaknesses of existing time series libraries, this paper introduces \textbf{\mbox{Tsururu}} (Fig.~\ref{fig:dataflow}), the modular framework for both practitioners and researchers that makes combinable global/multivariate approaches, different forecasting strategies, and models. This flexibility connects state-of-the-art research with real-world business applications.

\section{Framework Design}
\label{sec:framework}
The Tsururu architecture (Fig.~\ref{fig:dataflow}) facilitates the comparison of configurations and the development of task-specific forecasting systems, incorporating strategies with support for exogenous variables.

\textbf{Multi-series prediction strategies.}
Tsururu supports both \textbf{Global} and \textbf{Multivariate} approaches for all models for handling multiple time series \cite{januschowski2020criteria,montero2021principles}. The Global approach fits a single model to all time series, treating them as independent, while the Multivariate approach allows the model to capture dependencies between them. 
Moreover, each deep learning model in Tsururu supports \textbf{Channel Independence (CI)} and \textbf{Channel Mixing (CM)} modes \cite{han2024capacity}. The last one controls the interaction between series in the multivariate setting. 

\textbf{Multi-step-ahead prediction strategies.}
Different strategies can be applied to multi-step forecasting. 
\textbf{Recursive (Rec)} \cite{weigend2018time} trains a single model to predict the next point ($MH$ = 1), iteratively extending predictions across the forecast horizon and using previous predictions to update the features in the test data. $MH$ denotes the model horizon, i.e., the number of points which the model outputs in a single step.
Tsururu also supports the hybrid \textbf{Recursive-MIMO (Rec-MIMO)} strategy ($MH > 1$), which follows the recursive strategy but generates multiple-step predictions at each iteration instead of a single point.
\textbf{Direct (Dir)} \cite{weigend2018time} uses separate models for each prediction step with model horizon length, constructing the full forecast horizon. 
\textbf{MIMO} \cite{bontempi2008long} trains a single model to simultaneously predict the entire forecast horizon ($MH = H$).  
\textbf{FlatWideMIMO (FWM)} uses a single model to predict a specific point in the forecasting vector, with the horizon index explicitly provided as an input feature. While rarely used, this strategy can be effective.

\textbf{Pipeline and Data Transformations.}
Tsururu's pipeline applies sequential transformations to time series data, categorized into three types:
\textbf{Series-to-Series} are used for data preprocessing and feature generation, \textbf{Series-to-Features} build a ``wide'' series matrix with lagged versions of generated features, \textbf{Features-to-Features} perform window-based processing, such as LastKnownNormalizer (LKN), based on normalizing values by the most recent observed one in available history, i.e. in corresponding row of the “wide” series matrix. Tsururu supports separate transformations for features and targets, allowing more data preparation flexibility.

\textbf{Models.}
Tsururu offers classical ML and deep learning models. For ML, it includes boosting methods like CatBoost \cite{prokhorenkova2018catboost} and SketchBoost \cite{iosipoi2022sketchboost}, with SketchBoost chosen for its speed in GBDT training \cite{friedman2001greedy}. DL models are grouped into linear (DLinear \cite{zeng2023transformers}, CycleNet \cite{lincyclenet}), CNN-based (TimesNet \cite{wu2023timesnet}), and Transformer-based (PatchTST \cite{nietime}, GPT4TS \cite{zhou2023one}).

\textbf{Data flow.}
In Tsururu, data moves through a structured pipeline. First, time series pass through \textbf{preprocessing}.
Next, some \textbf{relevant features are generated}: categorical encodings and datetime features. Once these features are in place, Tsururu operates at the index level, converting ``long'' into ``wide'' series through lag transformations. This protocol prevents leakage from future values, supports dynamic feature generation, and reduces memory overhead. Afterward, Tsururu applies the beforehand chosen multi-step forecasting \textbf{strategy} and selects either the Global or Multivariate approach. \textbf{Window normalization} can then be used to rescale each observation in wide data to the most recent known value, thus helping to cope with local shifts in the data distribution. Finally, the prepared dataset and model are passed to the \textbf{Trainer} module. The Trainer is responsible for the training process. It also uses \textbf{cross validator (CV)} to generate validation splits. When multiple splits exist, a separate model is trained for each split, and their predictions are averaged during inference. The validation set can also be used for \textbf{early stopping}. The final model is then validated using backtesting and rolling validation methods to ensure robustness (note that it is not the same as CV in Trainer).
\section{Experimental Results}
\label{section:case}

\textbf{Setup.}
We examine such models as SketchBoost, DLinear, PatchTST, GPT4TS, and CycleNet, exploring both global and multivariate approaches (with either Channel Independence or Channel Mixing) and forecasting strategies: Recursive (with model horizon $MH$ equal to 1 or 6), MIMO, and FlatWideMIMO. These models are evaluated on the ILI dataset\footnote{\url{https://gis.cdc.gov/grasp/fluview/fluportaldashboard.html}}, a challenging weekly time series with annual periodicity and distinct temporal structures, to showcase the library's capabilities and highlight the importance of non-default model-approach-strategy combinations.

For all settings, we used a cosine-based scheduler for over 50 epochs, with the learning rate set to 0.0001, and performed parameter updates at the end of each batch. We fixed batch size to 32, history to 96, and horizon to 24.
The model hyperparameters (i.e., hidden dimension, the number of attention heads, etc.) were left unchanged from their earlier configurations in the initial works \cite{zhou2023one,zeng2023transformers}. In cases where the authors had not tested their models on ILI, we retained the hyperparameters for ETTh1 \cite{zhou2021informer}.

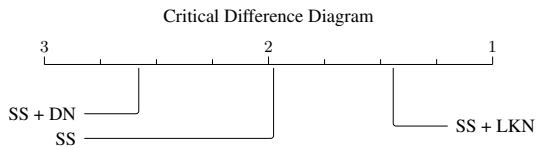
\begin{figure}[t]
\centering
\resizebox{0.85\linewidth}{!}{
\begin{tikzpicture}[
  treatment line/.style={rounded corners=1.5pt, line cap=round, shorten >=1pt},
  treatment label/.style={font=\normalsize},
  group line/.style={ultra thick},
]

\begin{axis}[
  clip={false},
  axis x line={center},
  axis y line={none},
  axis line style={-},
  xmin={1},
  ymax={0},
  scale only axis={true},
  width={\axisdefaultwidth},
  ticklabel style={anchor=south, yshift=1.3*\pgfkeysvalueof{/pgfplots/major tick length}, font=\small},
  every tick/.style={draw=black},
  major tick style={yshift=.5*\pgfkeysvalueof{/pgfplots/major tick length}},
  minor tick style={yshift=.5*\pgfkeysvalueof{/pgfplots/minor tick length}},
  title style={yshift=\baselineskip},
  xmax={3},
  ymin={-2.5},
  height={3\baselineskip},
  xtick={1,2,3},
  minor x tick num={3},
  x dir={reverse},
  title={Critical Difference Diagram},
]

\draw[treatment line] ([yshift=-2pt] axis cs:1.443609022556391, 0) |- (axis cs:1.193609022556391, -2.5)
  node[treatment label, anchor=west] {SS + LKN};
\draw[treatment line] ([yshift=-2pt] axis cs:1.9774436090225564, 0) |- (axis cs:2.8289473684210527, -3.0)
  node[treatment label, anchor=east] {SS};
\draw[treatment line] ([yshift=-2pt] axis cs:2.5789473684210527, 0) |- (axis cs:2.8289473684210527, -2.0)
  node[treatment label, anchor=east] {SS + DN};

\end{axis}
\end{tikzpicture}
}
\caption{A critical difference diagram visualizes the ranking of preprocessing methods across other fixed hyperparameters of the pipeline. Methods not connected by a horizontal line are significantly different. Here, we consider StandardScaler (SS) with LastKnownNormalizer (LKN) or DifferenceNormalizer (DN)}
\label{fig:preprocessing}
\end{figure}

\textbf{Results.} As shown in Figure \ref{fig:preprocessing}, LKN significantly outperforms default preprocessing strategies, demonstrating its effectiveness despite not being adopted in common libraries. We use delta-mode normalization (based on subtraction from the current value, the previous one for DifferenceNormalizer (DN), and the most recent one in available history for LKN).

\begin{table}[t]
\centering
\resizebox{\columnwidth}{!}{
\begin{tabular}{l|c|cc|cc|cc}
\toprule
\multirow{2}{*}{\textbf{Hyperparam}} & \multirow{2}{*}{\textbf{Value}} & \multicolumn{2}{c|}{\textbf{NN models}} & \multicolumn{2}{c|}{\textbf{Boosting}} & \multicolumn{2}{c}{\textbf{Overall}} \\
 & & \textbf{Rank} & \textbf{Median MAE} & \textbf{Rank} & \textbf{Median MAE} & \textbf{Rank} & \textbf{Median MAE} \\
\midrule
Datetime  & False & 1.3819 & 1.0087 & 1.3333 & 1.6050 & 1.3743 & 1.0448 \\
 Features& True & 1.6181 & 1.1323 & 1.6667 & 1.6174 & 1.6257 & 1.1785 \\
\midrule
\multirow[c]{2}{*}{ID Features} & False & 1.7262 & 1.0780 & 1.5714 & 1.6174 & 1.6952 & 1.1319 \\
 & True & 1.2738 & 1.0024 & 1.4286 & 1.5898 & 1.3048 & 1.0611 \\
\midrule
\multirow[l]{3}{*}{Mode} & Global & 1.5476 & 1.0056 & 1.0952 & 1.5648 & 1.5476 & 1.0735 \\
 & Multivariate CI & 2.2619 & 1.1217 & NaN & NaN & 2.2619 & 1.1217 \\
 & Multivariate CM & 2.1905 & 1.1319 & 1.9048 & 1.6248 & 2.1905 & 1.2129 \\
\midrule
& FlatWideMIMO & 3.9375 & 1.3080 & 2.8889 & 1.6208 & 3.7719 & 1.3543 \\
Prediction & MIMO & 1.7500 & 1.0280 & 2.4444 & 1.6072 & 1.8596 & 1.0621 \\
Strategy & Recursive ($MH=1$) & 2.4167 & 1.0314 & 2.7778 & 1.6066 & 2.4737 & 1.0763 \\
 & Recursive ($MH=6$) & 1.8958 & 1.0228 & 1.8889 & 1.5816 & 1.8947 & 1.0541 \\
\bottomrule
\end{tabular}
}
\caption{Comparison of hyperparameters of data manipulation pipeline. For boosting, there is no Multivariate CI mode by construction; only Multivariate CM mode is available. $MH$ stands for the model horizon of the Recursive strategy.}
\label{tab:params}
\end{table}


Table \ref{tab:params} presents the results of an ablation study on hyperparameters of the data manipulation pipeline, analyzing the impact of date and id features inclusion, forecasting mode, and strategy selection on model performance. The id features improved model accuracy, while including date features led to worse performance for both neural networks and GBDT. The Global approach outperformed the multivariate one, achieving the lowest rank and median MAE across all models. The MIMO strategy ranked best for neural networks, while Rec-MIMO ($MH=6$, see Section~\ref{sec:framework}) achieved the lowest median MAE. For GBDT, Rec-MIMO ($MH=6$) is the best strategy in rankings and median MAE. Thus, this hybrid approach is rarely used but is a competitive alternative.

\begin{table}[t]
\centering
\resizebox{\columnwidth}{!}{
\begin{tabular}{c|ccc|ccc}
\toprule
rank & Model & Strategy & MAE (test) & Model & Strategy & MAE (val) \\
\midrule
1 & GPT4TS & Recursive ($MH=6$) & 0.7804 & GPT4TS & MIMO & 0.2713 \\
2 & GPT4TS & Recursive ($MH=1$) & 0.7822 & GPT4TS & Recursive ($MH=6$) & 0.2833 \\
3 & PyBoost & FlatWideMIMO & 0.7921 & GPT4TS & Recursive ($MH=1$) & 0.2938 \\
4 & GPT4TS & MIMO & 0.7926 & PatchTST & MIMO & 0.3005 \\
5 & PatchTST & Recursive ($MH=6$) & 0.8630 & PatchTST & Recursive ($MH=6$) & 0.3050 \\
6 & PatchTST & MIMO & 0.8769 & DLinear & Recursive ($MH=6$) & 0.3169 \\
7 & PatchTST & Recursive ($MH=1$) & 0.8949 & PatchTST & Recursive ($MH=1$) & 0.3180 \\
8 & DLinear & Recursive ($MH=6$) & 0.9193 & DLinear & MIMO & 0.3205 \\
9 & DLinear & MIMO & 0.9220 & PyBoost & FlatWideMIMO & 0.3239 \\
10 & DLinear & Recursive ($MH=1$) & 0.9314 & DLinear & Recursive ($MH=1$) & 0.3313 \\
\bottomrule
\end{tabular}
}
\caption{Best 10 combinations model-strategy based on MAE on validation and test subsets.}
\label{tab:models}
\end{table}

Table \ref{tab:models} provides the independent ranking of models based on test and validation MAE. In this evaluation, we also considered an alternative scaling approach using ratio normalization without an initial StandardScaler for boosting models. Our results confirm that GPT4TS outperformed all other models. Notably, the Rec-MIMO strategy with $MH=6$ achieved the best overall test MAE. However, on the validation set, the highest-ranked strategy was MIMO. Interestingly, FlatWideMIMO combined with boosting models also ranked among the top strategies, demonstrating that GBDT can be competitive when paired with non-standard multi-step-ahead forecasting approaches. The diversity of top-ranked models and strategies underscores the importance of exploring rarely used model-strategy combinations. 

\textbf{Reproducibility.} To validate the accuracy of our time series library, we reproduce results from previous studies using Tsururu and compare them to the metrics reported in the original papers (Table \ref{tab:comparison}). The close alignment between these metrics demonstrates the fidelity of our implementations.

\begin{table}[t]
    \centering
    \tiny
    \resizebox{\columnwidth}{!}{
        \begin{tabular}{l|cc|cc}
        \toprule
        Model & MAE (Original) & MSE (Original) & MAE (Ours) & MSE (Ours) \\
        \midrule
        DLinear & 1.081 & 2.215 & 1.037 \tiny{$\pm$ 0.004} & 2.227 \tiny{$\pm$ 0.010} \\
        PatchTST & 0.754 & 1.319 & 0.716 \tiny{$\pm$ 0.043} & 1.239 \tiny{$\pm$ 0.103} \\
        GPT4TS & 0.881 & 2.063 & 0.896 \tiny{$\pm$ 0.024} & 2.028 \tiny{$\pm$ 0.054} \\
        TimesNet & 0.934 & 2.317 & 0.927 \tiny{$\pm$ 0.031} & 2.033 \tiny{$\pm$ 0.155} \\
        CycleNet & 1.073* & 2.400* & 1.051 \tiny{$\pm$ 0.013} & 2.345 \tiny{$\pm$ 0.049} \\
        \bottomrule
        \end{tabular}
    }
    \caption{Original vs. Ours.
    Metrics for the ILI dataset with a forecasting horizon of 24.
    Values marked with * indicate that the corresponding metric was not taken from the original paper but computed using the official implementation.
    Results are presented as mean ± standard deviation, calculated across three seeds.}
    \label{tab:comparison}
\end{table}

\section{Conclusion}

This paper introduces Tsururu, an open-source Python library for ablating all-with-all combinations of preprocessing, time series models, forecasting approaches, and strategies. It can be easily integrated with new SoTA models for fair benchmarking while providing key industrial features, such as exogenous variables and a Global approach to handling non-aligned time series. It also supports channel mixing and channel-independent forecasting, rarely used in existing libraries. Our experiments show the advantages of using rarely employed preprocessing (like LastKnownNormalizer) and combining strategies (like Recursive for PatchTST). Moreover, combining a Recursive strategy with the Global approach makes short time-series forecasting feasible, with extended horizons and support for custom datasets. The extended results for other datasets can be found in our repository. 
An ability to handle non-aligned series represents our essential advantage of combining regimes and strategies.

Future work includes incorporating Rectify \cite{taieb2014machine}, and DirRec \cite{sorjamaa2006time}, building a universal neural network constructor, testing patching techniques, and supporting time series with mixed discretization (daily, monthly, weekly, etc.) within multivariate datasets.


\section*{Ethical Impact}

The framework’s methods and specifications do not directly relate to ethical concerns. However, high-stakes domains such as healthcare and finance, where time series are widespread, require careful attention. Before deploying our library in such contexts, a thorough evaluation is essential to ensure it does not support decisions that could negatively impact individuals or organizations.

\section*{Acknowledgments}
The work of A. Savchenko was supported by a Research Center for Trusted Artificial Intelligence the Ivannikov Institute for System Programming of the Russian Academy of Sciences.

\bibliographystyle{named}
\bibliography{ijcai25}

\end{document}